\documentclass{egpubl}
\usepackage{pg2022}
 
\SpecialIssueSubmission    

\usepackage[T1]{fontenc}
\usepackage{dfadobe}  

\biberVersion
\BibtexOrBiblatex
\usepackage[backend=biber,bibstyle=EG,citestyle=alphabetic,backref=true]{biblatex} 
\AtEveryBibitem{%
\ifentrytype{article}{%
}{}
\ifentrytype{inproceedings}{%
  \clearfield{volume}
  \clearfield{number}
}{}
  \clearlist{publisher}
  \clearfield{isbn}
  \clearlist{location}
  \clearfield{doi}
  \clearfield{url}
}
\electronicVersion
\PrintedOrElectronic

\ifpdf \usepackage[pdftex]{graphicx} \pdfcompresslevel=9
\else \usepackage[dvips]{graphicx} \fi

\usepackage{egweblnk} 

\usepackage{amsmath}
\usepackage{amssymb}
\usepackage{algorithm}
\usepackage{algorithmic}
\usepackage{mathbbol}
\usepackage{multirow}
\newcommand{\best}[1]{\textbf{#1}}
\newcommand{\sbest}[1]{#1}
\usepackage{color}
\newcommand{\chkA}[1]{#1}

\title[
User-Controllable Latent Transformer for StyleGAN Image Layout Editing
]%
      {
      User-Controllable Latent Transformer \\for StyleGAN Image Layout Editing
      }

\author[Y. Endo]
{\parbox{\textwidth}{\centering Y. Endo\orcid{0000-0001-5132-3350}
        }
        \\
{\parbox{\textwidth}{\centering University of Tsukuba, Japan
       }
}
}

\volume{41}   

\begin{document}

\teaser{
\vspace{-4.5mm}
 \includegraphics[width=\linewidth]{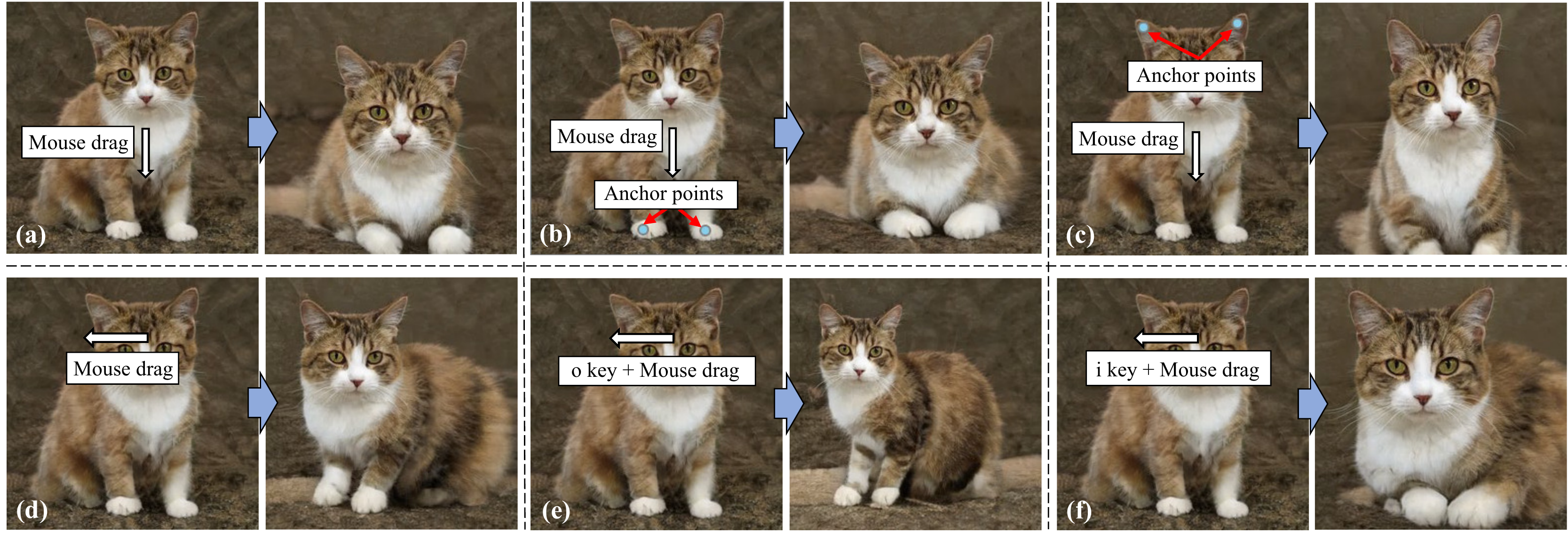}
 \centering
 \vspace{-4mm}\caption{
Editing StyleGAN image layout using our user-controllable latent transformer. As shown in (a) and (d), our method can interactively generate an image reflecting a user-specified movement direction (white arrows) via manipulation in a latent space. As shown in (b) and (c), the user can specify the locations where the user does not want to move with anchor points (blue circles). 
Our method can also handle 3D motion with an additional key input (denoted as the ``o'' or ``i'' key), as shown in (e) and (f). 
  }
  \label{fig:teaser}
}

\maketitle
\begin{abstract}
Latent space exploration is a technique that discovers interpretable latent directions and manipulates latent codes to edit various attributes in images generated by generative adversarial networks (GANs). 
However, in previous work, spatial control is limited to simple transformations (e.g., translation and rotation), and it is laborious to identify appropriate latent directions and adjust their parameters. 
In this paper, we tackle the problem of editing the StyleGAN image layout by annotating the image directly. 
To do so, we propose an interactive framework for manipulating latent codes in accordance with the user inputs. 
In our framework, the user annotates a StyleGAN image with locations they want to move or not and specifies a movement direction by mouse dragging. 
From these user inputs and initial latent codes, our \textit{latent transformer} based on a transformer encoder-decoder architecture estimates the output latent codes, which are fed to the StyleGAN generator to obtain a result image. 
To train our latent transformer, we utilize synthetic data and pseudo-user inputs generated by off-the-shelf StyleGAN and optical flow models, without manual supervision. 
Quantitative and qualitative evaluations demonstrate the effectiveness of our method over existing methods. 


\end{abstract}  

\section{Introduction}
Generative adversarial networks (GANs) have attracted much attention due to their ability to generate photorealistic images and a wide range of applications~\cite{DBLP:conf/iclr/KarrasALL18,DBLP:conf/iclr/BrockDS19,DBLP:conf/cvpr/KarrasLA19,DBLP:conf/cvpr/KarrasLAHLA20,DBLP:conf/nips/KarrasALHHLA21}. 
GANs are generative models that learn real image distributions via adversarial learning and can generate diverse images from random vectors in a low-dimensional latent space. 
To edit GAN images by manipulating latent codes, latent space exploration techniques have been actively studied~\cite{DBLP:conf/icml/VoynovB20,DBLP:conf/cvpr/ShenGTZ20,gansteerability,DBLP:journals/corr/abs-2004-02546,DBLP:conf/cvpr/YangCWZSH21,DBLP:journals/corr/abs-2007-06600,DBLP:conf/iccv/YukselSEY21,DBLP:journals/tog/AbdalZMW21}. 
By moving latent codes toward specific directions found by these techniques, the user can edit various image attributes, such as facial orientation, pose, age, and gender. 

Existing latent space exploration techniques aim to discover interpretable directions in a latent space. 
Among latent directions found by these techniques, the user selects a latent direction corresponding to a specific attribute change and then adjusts the manipulation amount with a slider user interface (UI). 
However, such operations do not always lead to intuitive editing. 
In particular, spatial control for diverse layouts (e.g., pose and shape) is difficult to handle via 1D operations based on a slider UI. 
In addition, in most existing techniques, spatial control is limited to basic transformations, such as translation, rotation, and scaling. 
Even if we could discover latent directions for any layout control, identifying them and adjusting their parameters are laborious. 

In this paper, we tackle the novel problem of controlling the spatial layout of StyleGAN images by manipulating latent codes in accordance with user inputs directly specified on the images. 
Figure~\ref{fig:teaser} shows examples of this problem, where the user specifies several annotations on the StyleGAN images to edit them. 
As shown in (a), the user specifies a motion vector on the cat image by mouse dragging, with the aim to find a latent direction such that the clicked location moves to the specified direction. 
The system outputs editing results in real time by moving the initial latent codes toward the found directions as the mouse is dragged. 
However, in this case, the single annotation leaves ambiguity about whether other parts of the cat should move or not. 
Therefore, as shown in (b) and (c), we introduce an anchor point (AP) interface to enable the user to specify points that should not move. 
In addition to the 2D motion shown in (d), we consider a 3D motion by using additional key inputs, as shown in (e) and (f). 

To this end, we propose a framework for manipulating latent codes in accordance with user inputs specified on the StyleGAN images. 
In our framework, we introduce a \textit{latent transformer} that estimates latent directions conditioned on multiple user inputs. 
A latent transformer was recently presented by Yao et al.~\cite{Yao_2021_ICCV}, and it is based on a multilayer perceptron (MLP) to perform a latent code transformation for a specific facial attribute. 
\chkA{Patashnik et al.~\cite{DBLP:conf/iccv/PatashnikWSCL21} and} Khodadadeh et al.~\cite{DBLP:conf/wacv/KhodadadehGMLBK22} also proposed MLP-based latent \chkA{transformers} for multiple facial attributes \chkA{or text-driven manipulation. }
However, their methods are not suitable for our problem because they can only handle fixed-length inputs (e.g., a latent code and an attribute vector). 
We develop a latent transformer based on a transformer encoder-decoder architecture to handle variable-length user inputs. 
In addition to latent codes and user inputs, our latent transformer utilizes StyleGAN feature vectors as input, enabling us to consider the semantics of specified locations. 
Our latent transformer estimates latent directions using these inputs and moves the input latent codes to those directions.
We can obtain images reflecting the user inputs by feeding the manipulated latent codes to the StyleGAN generator. 
Our latent transformer is trained using synthetic data generated from randomly sampled and perturbed latent codes. 
We estimate forward flows from the two images generated from the pair of latent codes and treat them as pseudo-user inputs. 
Therefore, our method requires no additional training data and can be implemented with off-the-shelf StyleGAN and optical flow models. 

We conduct quantitative and qualitative evaluation experiments on various datasets to validate the effectiveness of our method. We demonstrate that our approach enables the interactive spatial control of StyleGANs in accordance with user inputs specified with our UI, without identifying latent directions and adjusting their parameters as previous work does. The main contributions of this paper are three-fold: (i) a framework for controlling StyleGAN image layout in accordance with user inputs on images; (ii) a latent transformer based on a transformer encoder-decoder; (iii) a training pipeline for our latent transformer using synthetic images and pseudo-user inputs, without manual supervision. 
\begin{figure*}[t]
  \centering
  \includegraphics*[width=1\linewidth]{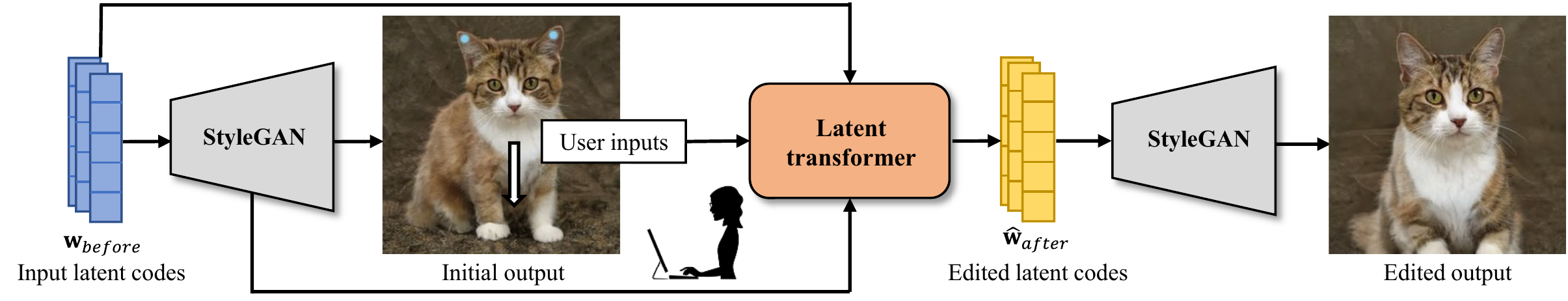}
  \vspace{-4mm}\caption{
Overview of our inference pipeline. The user directly annotates the StyleGAN image generated from the initial latent codes. The latent transformer computes output latent codes using the annotations, initial latent code, and StyleGAN feature map as input. The output latent codes are fed to StyleGAN to obtain the edited image. 
}
  \label{fig:inference}
\end{figure*}

\section{Related Work}
\subsection{Generative adversarial networks}
GANs are generative models based on a two-player game between a generator and discriminator. 
The progress of GANs has been remarkable in the past few years, with PGGAN~\cite{DBLP:conf/iclr/KarrasALL18}, BigGAN~\cite{DBLP:conf/iclr/BrockDS19}, and StyleGAN~\cite{DBLP:conf/cvpr/KarrasLA19,DBLP:conf/cvpr/KarrasLAHLA20}, which are able to generate photorealistic and high-resolution images. 
In particular, StyleGAN can give us control over different levels (i.e., coarse, middle, and fine) thanks to the network architecture that takes a latent code controlling each layer as input. 
For example, manipulating latent codes on deep layers enables spatial control, such as pose and orientation. 
To enable intuitive spatial control over StyleGAN, our work provides a framework that manipulates latent codes in accordance with user inputs directly specified on the image. 

\chkA{
More recent GANs introduce additional disentangled parameters to control output layouts. 
StyleNeRF~\cite{gu2021stylenerf} and StyleGAN3~\cite{DBLP:conf/nips/KarrasALHHLA21} use camera parameters or Fourier features as input besides latent codes. 
However, layout-related editing using these parameters is limited to camera poses (e.g., yaw and pitch) or affine transformation (e.g., translation and rotation). They must also identify interpretable latent directions in a latent space to edit more complex layouts. Furthermore, only using these models cannot edit layouts by annotating images directly. Our work does not compete against their work strongly but rather could improve the editability of latent manipulation for their work. 
StyleRig~\cite{DBLP:conf/cvpr/TewariEBBSPZT20} provides rig-like controls for StyleGAN via a parametric face model. This method specializes in facial images, while our approach is more general and can be applied to various datasets. 
}

Image-to-image translation based on conditional GANs has also been actively studied~\cite{DBLP:conf/cvpr/IsolaZZE17,DBLP:conf/iccv/ZhuPIE17,DBLP:conf/cvpr/Wang0ZTKC18,DBLP:conf/cvpr/Park0WZ19,DBLP:conf/cvpr/0004YLLZHH19}. 
These methods can also enable layout control using \chkA{intuitive user inputs such as semantic masks}~\cite{DBLP:journals/cvm/XueGZXZH22}. 
However, most methods require a large amount of training data containing manually created semantic masks. 
In contrast, our work does not require manual supervision. 

\subsection{Latent space exploration}
Latent space exploration techniques, which aim to discover meaningful directions in GANs' latent spaces, are mainly categorized into supervised and unsupervised approaches. 
For the supervised approach, InterFaceGAN~\cite{DBLP:conf/cvpr/ShenGTZ20} trained a support vector machine (SVM) for facial attribute classification in a latent space and achieved attribute editing by manipulating a latent code toward a normal direction of a hyperplane.  
StyleFlow~\cite{DBLP:journals/tog/AbdalZMW21} can also edit facial attributes using conditional continuous normalization flows. 
While these two methods require positive and negative examples to train classifiers, Yang et al.~\cite{DBLP:conf/cvpr/YangCWZSH21} proposed an attribute editing method that requires positive examples only. 
There also exist self-supervised approaches ~\cite{gansteerability,DBLP:conf/iclr/SpingarnBM21}.
For example, Jahanian et al. ~\cite{gansteerability} applied simple image transformations, such as zooming in and out, to source images and optimized the latent directions by minimizing the difference between the transformed source images and the GAN outputs. 
However, spatial control by these supervised and self-supervised approaches is limited because binary classifiers and simple image transformations cannot handle various layout changes. 

For the unsupervised approach~\cite{DBLP:conf/nips/ChenCDHSSA16,DBLP:conf/icml/VoynovB20,DBLP:journals/corr/abs-2004-02546,DBLP:journals/corr/abs-2007-06600,DBLP:conf/iccv/HeKS21,DBLP:conf/iccv/YukselSEY21,zhu2021lowrankgan}, for example, GANSpace~\cite{DBLP:journals/corr/abs-2004-02546} discovered that moving a latent code toward principal directions in a latent space leads to interpretable control.  
SeFa~\cite{DBLP:journals/corr/abs-2007-06600} is a closed-form method to find semantic latent directions by eigen decomposition of the network weights. 
LatentCLR~\cite{DBLP:conf/iccv/YukselSEY21} used contrastive learning to improve discriminability between latent directions. 
Although these unsupervised approaches can discover interpretable latent directions without supervision, 
spatial control is limited to basic operations, such as rotation, translation, and scaling. 
Furthermore, the user needs to manually identify latent directions and adjust their parameters. 
In contrast, our method can manipulate latent codes via intuitive annotation on images. 

\section{Method}
This study aims to intuitively control the output images of StyleGAN by allowing the user to specify motion vectors directly on the images. 
We formulate this task as the problem of constructing a latent transformer $\mathbf{T}$, which transforms initial latent codes $\mathbf{w}_{before}$ in accordance with user inputs $\mathcal{U}$:
\begin{align}
\mathbf{T}(\mathbf{w}_{before}, \mathcal{U}, \alpha) = \mathbf{w}_{before} + \alpha \cdot f(\mathbf{w}_{before}, \mathcal{U}), 
\label{eq:lt}
\end{align}
where $\alpha$ is a parameter that adjusts the degree of manipulation for the latent codes $\mathbf{w}_{before}$, and $f$ is an arbitrary function based on a neural network. 
The user inputs are defined as $\mathcal{U} = \{\mathbf{v}_i, \mathbf{p}_i\}_{i=1}^{K}$ consisting of $K$ motion vectors $\mathbf{v}_i \in \mathbb{R}^3$ in the $xyz$-directions and pixel positions $\mathbf{p}_i \in \mathbb{Z}^{2}$ of the start points for $\mathbf{v}_i$. 
Although several latent transformers for manipulating specific attributes have been proposed~\cite{Yao_2021_ICCV,DBLP:conf/iccv/PatashnikWSCL21,DBLP:conf/wacv/KhodadadehGMLBK22}, our work is the first attempt to construct the latent transformer conditioned on such user inputs. 

Through the overview of the inference pipeline in Figure ~\ref{fig:inference}, we explain the flow of interactive image editing using our latent transformer. 
First, the user annotates the output image of StyleGAN from the initial latent codes $\mathbf{w}_{before}$ (Section~\ref{sec:interface}).
Next, we inject the user input $\mathcal{U}$, initial latent codes $\mathbf{w}_{before}$, and StyleGAN feature map into the latent transformer to compute the edited latent codes $\hat{\mathbf{w}}_{after}$ (Section~\ref{sec:arch}). 
Finally, we obtain the resulting image by injecting the latent codes $\hat{\mathbf{w}}_{after}$ into the StyleGAN generator. 
There is no need to manually create a training dataset to train the latent transformer because we use synthetic images and pseudo-user inputs generated with pre-trained StyleGAN and optical flow network models (Section \ref{sec:training}).

\begin{figure*}[t]
  \centering
  \includegraphics*[width=1\linewidth]{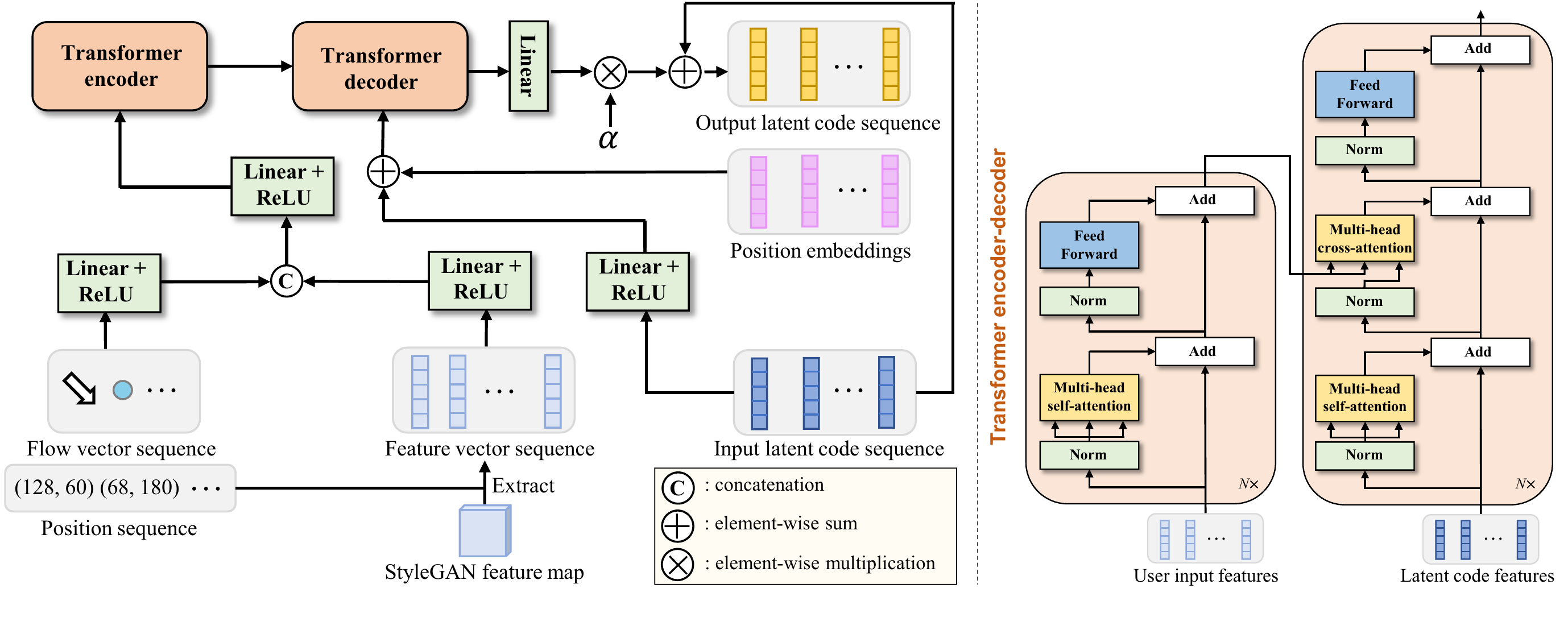}
  \vspace{-4mm}\caption{
Network architectures of our latent transformer (left) and transformer encoder-decoder (right). The transformer encoder takes a motion vector feature sequence and StyleGAN features extracted using a position sequence. The transformer decoder takes the output of the transformer encoder and a latent code feature sequence added with position embeddings. 
The transformer decoder outputs latent directions in accordance with the user inputs, scaled by $\alpha$ and added to the input latent code sequence. 
}
  \label{fig:transformer}
\end{figure*}

\subsection{Network architecture}
\label{sec:arch}
Figure~\ref{fig:transformer} shows the architecture of our latent transformer. 
To handle a different number of user inputs for each test time, we incorporate a transformer encoder-decoder architecture, which can handle variable-length inputs, in our latent transformer. 

On the side of the transformer encoder, given user inputs $\mathcal{U}$, it extracts a sequence of feature vectors passed to the transformer decoder. 
Because pixel positions $\mathbf{p}_i$ themselves do not contain semantic information about what is to be moved, we instead use semantic feature vectors extracted from the StyleGAN feature map as input. 
This idea comes from recent studies that use StyleGAN feature maps for semantic segmentation tasks~\cite{DBLP:conf/cvpr/CollinsBPS20,DBLP:conf/cvpr/ZhangLGYLB0F21,DBLP:conf/cvpr/TritrongRS21}. 
Specifically, we compute a $64\times64$ intermediate feature map from StyleGAN using $\mathbf{w}_{before}$ as input and then extract a sequence of StyleGAN feature vectors corresponding to pixel positions ${\mathbf{p}}_i$. 
We then merge these two inputs (i.e., motion vector sequence and StyleGAN feature vector sequence) to pass them to the transformer encoder. 
To do so, we convert them into sequences of 256-channel vectors through the linear layers. 
We concatenate them to obtain a sequence of 512-channel vectors and pass them through the linear layer. 
In the transformer encoder, the self-attention layer uses the obtained 512-channel vector sequence as a key, value, and query to extract global features capturing the relationship among multiple user inputs. 

On the side of the transformer decoder, given the output of the transformer encoder and the latent codes $\mathbf{w}_{before}$, it computes edited latent codes $\hat{\mathbf{w}}_{after}$. 
In our approach, we assume that the latent code for controlling each StyleGAN layer may differ. 
That is, the transformer decoder treats a sequence of 512-channel vectors as input and output latent codes (i.e., latent codes in the $\mathcal{W}^+$ space~\cite{DBLP:conf/iccv/AbdalQW19}). 
The input latent code sequence is passed through the linear layer and added with position embeddings. 
The position embeddings are learnable parameters and help distinguish the latent codes for each layer. 
The transformer decoder extracts features capturing the relationship between the user inputs and latent codes. In particular, the cross attention layer takes the transformer encoder output as a key and value and takes the sequence of feature vectors based on the latent codes as a query. 
Latent directions can then be computed from the transformer decoder output via the linear layer. 
We finally obtain $\hat{\mathbf{w}}_{after}$ by scaling the latent directions by $\alpha$ and adding them to the initial latent codes $\mathbf{w}_{before}$. 

\subsection{Training}
\label{sec:training}
\begin{figure*}[t]
  \centering
  \includegraphics*[width=1\linewidth]{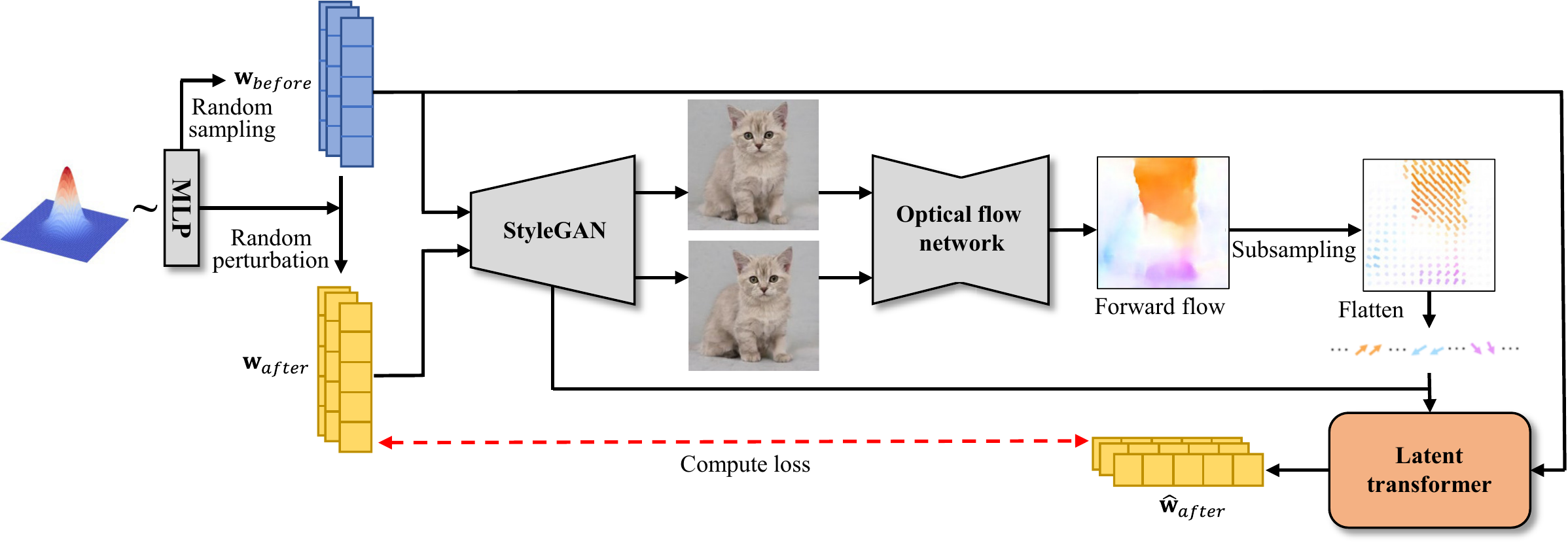}
  \vspace{-4mm}\caption{
Training pipeline of our latent transformer. We first sample initial latent codes from a normal distribution and further perturb them. Then, an optical flow network estimates a forward flow field between images obtained from these latent codes. From the sub-sampled forward flows, initial latent codes, and StyleGAN feature map, our latent transformer estimates the edited latent codes. Finally, we minimize the loss between the perturbed latent code and estimated one and update the weights of the latent transformer by backpropagation. 
}
  \label{fig:overview}
\end{figure*}

We train our latent transformer previously described using synthetic images from StyleGAN. 
Figure ~\ref{fig:overview} illustrates the flow of a training iteration. 
First, we randomly sample the latent codes $\mathbf{w}_{before}$ from the prior. 
To reduce artifacts in the output image from $\mathbf{w}_{before}$, we use the truncation trick~\cite{DBLP:conf/cvpr/KarrasLA19}, which is linear interpolation using the average latent code $\bar{\mathbf{w}}$:
\begin{align}
\mathbf{w}_{before} = \bar{\mathbf{w}} - \psi (\mathbf{w}_{rand}-\bar{\mathbf{w}}),
\end{align}
where $\psi$ is a constant and $\mathbf{w}_{rand}$ is a random latent code obtained via the StyleGAN mapping network from Gaussian noise. 
Next, to slightly change the layout of the image generated from $\mathbf{w}_{before}$, we randomly perturb it as follows:
\begin{align}
    \mathbf{w}_{after} = \mathbf{w}_{before} - \phi (\mathbf{w}'_{rand}-\mathbf{w}_{before}),
\end{align}
where $\phi$ is a constant and $\mathbf{w}'_{rand}$ is another random latent code from prior. 
For $\mathbf{w}'_{rand}$, we sample a different vector for controlling each StyleGAN layer to handle various layout changes. 
For StyleGAN, it is known that the latent codes for the deep layers affect coarse styles (e.g., pose and shape), while those for the shallow layers affect fine styles (e.g., color and texture). We therefore use the six latent codes controlling the deepest layers to train the latent transformer. 

For the two output images from $\mathbf{w}_{before}$ and $\mathbf{w}_{after}$, we compute a forward flow field using a pre-trained optical flow network. 
We use this flow field as pseudo-user inputs in the training time. 
To handle 3D motion, we use Yang and Ramanan's method~\cite{DBLP:conf/cvpr/YangR20}, which can estimate optical flow offering the position change and optical expansion offering the scale change. 
We compute a 3D motion vector $(\frac{x_j}{\sigma_f}, \frac{y_j}{\sigma_f}, \frac{z_j}{\sigma_e})^T$ for each pixel $j$ using optical flow $(x_j, y_j)$ and optical expansion $z_j$. 
Here, $\sigma_f$ and $\sigma_e$ are constants for normalizing optical flow and optical expansion, respectively, which differ significantly in range. 
To compute these two parameters, we randomly sample hundreds of $\mathbf{w}_{before}$ and $\mathbf{w}_{after}$ pairs and estimate optical flow and optical expansion maps for each pair. We define $\sigma_f$ and $\sigma_e$ as the averages of the maximum values of these optical flow and optical expansion maps, respectively. 

Next, we feed the flow field to the latent transformer. 
However, it is computationally challenging for the transformer encoder-decoder network to handle a sequence of flows for all pixels. 
Hence, we subsample the flow field and use $16 \times 16$ one as input. 
In addition to the flow field, we feed $\mathbf{w}_{before}$ and a StyleGAN feature map to the latent transformer and obtain the output latent codes $\hat{\mathbf{w}}_{after}$. 
We set $\alpha$ to 1 during the training time. 
Finally, we update the weights of the latent transformer by minimizing the L2 loss between $\hat{\mathbf{w}}_{after}$ and $\mathbf{w}_{after}$. 
By iterating the aforementioned cycle, the latent transformer can learn to transform latent codes in accordance with various flow fields for synthetic images. 

\subsection{User interface}
\label{sec:interface}
In this section, we first briefly describe the UI for specifying user inputs~$\mathcal{U}$ fed to the latent transformer during the test time.
As mentioned in Section~\ref{sec:training}, we train our latent transformer on pseudo-user inputs, but manually providing similar dense flows is impractical. 
Fortunately, our latent transformer can handle variable-length inputs, and thus we adopt an interface to specify a single motion vector and multiple APs, as shown in Figure~\ref{fig:teaser}(a)--(c). 
The user first specifies locations not to be moved with APs and then specifies a location to be moved and movement direction by mouse dragging. 
As the mouse is dragged, the user can get an inference result instantly from the system. 
Because mouse dragging can only handle 2D motion, we use the ``i'' or ``o'' key for $z$-directional operations at the same time, as shown in Figure ~\ref{fig:teaser}(d)--(f). 
Simply zooming in or out can also be achieved by the mouse wheel. 

Next, we describe how to feed user inputs specified with the aforementioned UI to the latent transformer. 
For mouse dragging, we compute a motion vector $\mathbf{v}_i \in \mathcal{U}$ with the start point $\mathbf{s}_i \in \mathbb{Z}^2$ and the end point $\mathbf{e}_i \in \mathbb{Z}^2$ of the mouse as follows: 
\begin{align}
\mathbf{v}_i = \left( \frac{(\mathbf{e}_i-\mathbf{s}_i)^T}{\|\mathbf{e}_i-\mathbf{s}_i\|}, 0 \right)^T. 
\label{eq:mouse}
\end{align}
In this equation, we normalize the motion vector to avoid using values that are extremely far from the distribution of the training data. 
However, this normalization prevents $\mathbf{v}_i$ from determining the amount of movement, and thus we control $\alpha$ in Equation (\ref{eq:lt}) instead: 
\begin{align}
\alpha = \beta \|\mathbf{e}_i-\mathbf{s}_i\|, 
\end{align}
where $\beta$ is a constant that determines how much effect user inputs have on latent code manipulation. 
We also set the pixel position $\mathbf{p}_i \in \mathcal{U}$ to the start point $\mathbf{s}_i$. 
To further specify motion in the $z$-direction, we assign a constant to the $z$-coordinate of $\mathbf{v}_i$ in accordance with keystrokes or mount of mouse wheel rotation. 
In our experiments, we assign -5 to the $z$-coordinate of $\mathbf{v}_i$ for zooming in and 5 for zooming out.
When simply zooming in and out using the mouse wheel without specifying a 2D motion, we add a constant to $\alpha$ in accordance with the amount of mouse wheel rotation.
For the APs, we set $\mathbf{v}_i$ to a zero vector and $\mathbf{p}_i$ to the positions of the APs. 

\begin{table*}[]
\begin{center}
\caption{
Quantitative comparison. The value after ``Ours-'' denotes the number of user inputs $K$. The bold font indicates the best score for each metric in each dataset. 
}
\label{tab:quantitative_res}
\small
\begin{tabular}{l|l|cccccc}
Dataset & \multicolumn{1}{l|}{Metric}                                            & \begin{tabular}[c]{@{}c@{}}SeFa~\cite{DBLP:journals/corr/abs-2007-06600}\\ (random)\end{tabular} & \begin{tabular}[c]{@{}c@{}}SeFa~\cite{DBLP:journals/corr/abs-2007-06600}\\ (greedy)\end{tabular} & \begin{tabular}[c]{@{}c@{}}LatentCLR~\cite{DBLP:conf/iccv/YukselSEY21}\\ (random)\end{tabular} & \begin{tabular}[c]{@{}c@{}}LatentCLR~\cite{DBLP:conf/iccv/YukselSEY21}\\ (greedy)\end{tabular} & Ours-1                   & Ours-32                                         \\ \hline
    & MSE $\times 10^2 \downarrow$ & 5.69                                                    & \sbest{1.54} & 7.34                                                    & 3.09                          & 1.84                     & \best{1.26}                     \\
cat                        & LPIPS $\times 10 \downarrow$ & 2.76                                                    & \sbest{0.98} & 3.04                                                    & 1.79                         & 1.12                     & \best{0.89}                     \\
                        & FID $\downarrow$                                                       & 14.27                                                   & \sbest{7.02}  & 16.37                                                   & 9.17                         & 7.14                     & \best{6.47}                     \\ \hline
 & MSE $\times 10^2 \downarrow$ & 6.99                                                    & \sbest{2.46} & 8.47                                                    & 6.07                          & 2.99                     & \best{2.41}                     \\
church                        & LPIPS $\times 10 \downarrow$  & 2.21                                                    & \best{0.98} & 2.27                                                    & 1.66                           & 1.18                     & \sbest{1.03}                    \\
                        & FID $\downarrow$                                                       & 6.51                                                    & \sbest{4.57}  & 7.16                                                    & 5.74                           & 4.90                     & \best{4.51}                     \\ \hline
    & MSE $\times 10^2 \downarrow$ & 8.40                                                    & \sbest{2.31}  & 9.31                                                    & 4.11                         & 2.87                     & \best{1.96}                     \\
car                        & LPIPS $\times 10 \downarrow$  & 2.65                                                    & \sbest{0.92} & 2.86                                                    & 1.56                          & 1.11                     & \best{0.85}                     \\
                        & FID $\downarrow$                                                       & 19.26                                                   & \sbest{7.68} & 22.37                                                   & 9.38                          & 8.28                     & \best{7.38}                     \\ \hline
   & MSE $\times 10^2 \downarrow$ & 10.28                                                   & \sbest{0.87} & 16.64                                                   & 1.87                           & 1.33                     & \best{0.74}                    \\
ffhq                        & LPIPS $\times 10 \downarrow$  & 3.70                                                    & \sbest{0.76} & 5.07                                                    & 1.81                          & 0.97                     & \best{0.67}                   \\
                        & FID $\downarrow$                                                       & 32.49                                                   & \sbest{3.98}  & 123.57                                                  & 6.65                           & 4.45                     & \best{3.65}                    \\ \hline
  & MSE $\times 10^2 \downarrow$ & {14.41}                               & \sbest{3.58} & 14.28                               & 4.08      & 5.15 & \best{3.09} \\
anime                        & LPIPS $\times 10 \downarrow$  & {4.21}                                & \sbest{1.27}  & 4.18                                & 1.54      & 1.62 & \best{1.09} \\
                        & FID $\downarrow$                                                       & {25.72}                               & \sbest{6.56}  & 41.83                               & 7.39      & 7.10 & \best{6.01} \\ \hline
\end{tabular}
\end{center}
\end{table*}

\section{Experiments}
\subsection{Implementation details}
We implemented our method using PyTorch and ran our program on a PC equipped with RTX A4000 GPUs.
We used the StyleGAN2 generators pre-trained with the cat, church, car, ffhq, and anime portrait datasets~\cite{awesomeStyleGAN2}. 
For the transformer encoder in our latent transformer, we used the same transformer encoder architecture as the vision transformer (ViT)~\cite{DBLP:conf/iclr/DosovitskiyB0WZ21}. 
For the transformer decoder, we based it on the original architecture~\cite{DBLP:conf/nips/VaswaniSPUJGKP17} but adopted PreNorm~\cite{ DBLP:conf/acl/WangLXZLWC19}, which applies normalization before the sublayers, and replaced the ReLU function with the GeLU function~\cite{hendrycks2016gelu} in the feed-forward layers. 
For each transformer encoder and decoder, we set the number of multi-head attention heads to 8 and the number of layers ($N$ in Figure~\ref{fig:transformer}) to 6. 
To train the latent transformer, we used the Ranger optimizer~\cite{Ranger} with a learning rate of 0.001 and set the parameters $\psi$ and $\phi$ to $0.3$ and $0.1$, respectively. 
We trained the latent transformer for 60,000 iterations with a batch size of 1. 
Training took about 4 hours for $256\times256$ images and about 7 hours for $1024\times1024$ images, and inference took about 0.02 and 0.07 seconds for each image size. 

\subsection{Quantitative evaluation}
\label{sec:eval_quantitative}
\paragraph*{Datasets.} 
In this section, we evaluate the effectiveness of our method quantitatively using synthetic datasets, which were prepared as follows. 
First, we randomly sampled latent codes $\mathbf{w}_{before}$ and $\mathbf{w}_{after}$ using different seeds from the training time. 
Next, we extracted a flow field from the pair of the StyleGAN images from $\mathbf{w}_{before}$ and $\mathbf{w}_{after}$ using the pre-trained optical flow network.
Finally, we randomly sampled $K$ pixels from the flow field and used them as the user inputs $\mathcal{U}$. 
We generated 1,000 triplets ($\mathbf{w}_{before}$, $\mathcal{U}$, $\mathbf{w}_{after}$) for each pre-trained StyleGAN model. 
For evaluation, we computed evaluation metrics between images from the perturbed latent codes $\mathbf{w}_{after}$ and the estimated ones and averaged them for 1,000 samples. 
As evaluation metrics, we used MSE, LPIPS~\cite{DBLP:conf/cvpr/ZhangIESW18}, and Clean-FID~\cite{parmar2021cleanfid}. 

\paragraph*{Compared methods.} 
To the best of our knowledge, our work is the first to use motion vectors as input to control StyleGAN, and no prior work for tackling the similar problem exists. 
Possible candidates for comparison would be latent space exploration methods, mainly categorized into supervised and unsupervised approaches. 
Although the supervised approach~\cite{DBLP:conf/cvpr/ShenGTZ20,DBLP:journals/tog/AbdalZMW21,DBLP:conf/cvpr/YangCWZSH21} requires annotated images or attribute classifiers to find interpretable latent directions, defining various layouts as explicit class labels is difficult. 
The unsupervised approach can find latent directions via latent space analysis on the basis of eigenvalue decomposition~\cite{DBLP:journals/corr/abs-2007-06600} and contrastive learning~\cite{DBLP:conf/iccv/YukselSEY21}, without additional data and classifiers.
We therefore used the state-of-the-art unsupervised methods, SeFa~\cite{DBLP:journals/corr/abs-2007-06600} and LatentCLR~\cite{DBLP:conf/iccv/YukselSEY21} for comparison. 

Unlike our method, however, the unsupervised methods are not designed for associating latent code directions with the user inputs $\mathcal{U}$. 
In these methods, the user needs to identify what attribute each latent direction affects and adjust parameters for latent directions to obtain desired images. 
To do so automatically, we designed two approaches (\textit{random} and \textit{greedy}) for the existing methods. 
The random approach randomly manipulates the input latent codes in a certain manipulation range along with $k$ latent directions a hundred times. 
It then selects the output image with the smallest LPIPS for the ground-truth image. 
In addition, among $k$ latent directions, the greedy approach greedily searches for a single latent direction and parameter that generate the image with the smallest LPIPS. 
In other word, the greedy approach simulates the user that searches for the best latent direction and parameter for reproducing a ground-truth image. 
On the basis of their official codes and experiments, we set $k=50$ and the manipulation range to $[-3,+3]$ for SeFa and $k=100$ and the manipulation range to $[-15,+15]$ for LatentCLR. 
For the greedy approach, we searched for parameters by dividing the manipulation range into 11 values for each latent direction.

\begin{figure}[t]
  \centering
  \includegraphics*[width=1\linewidth]{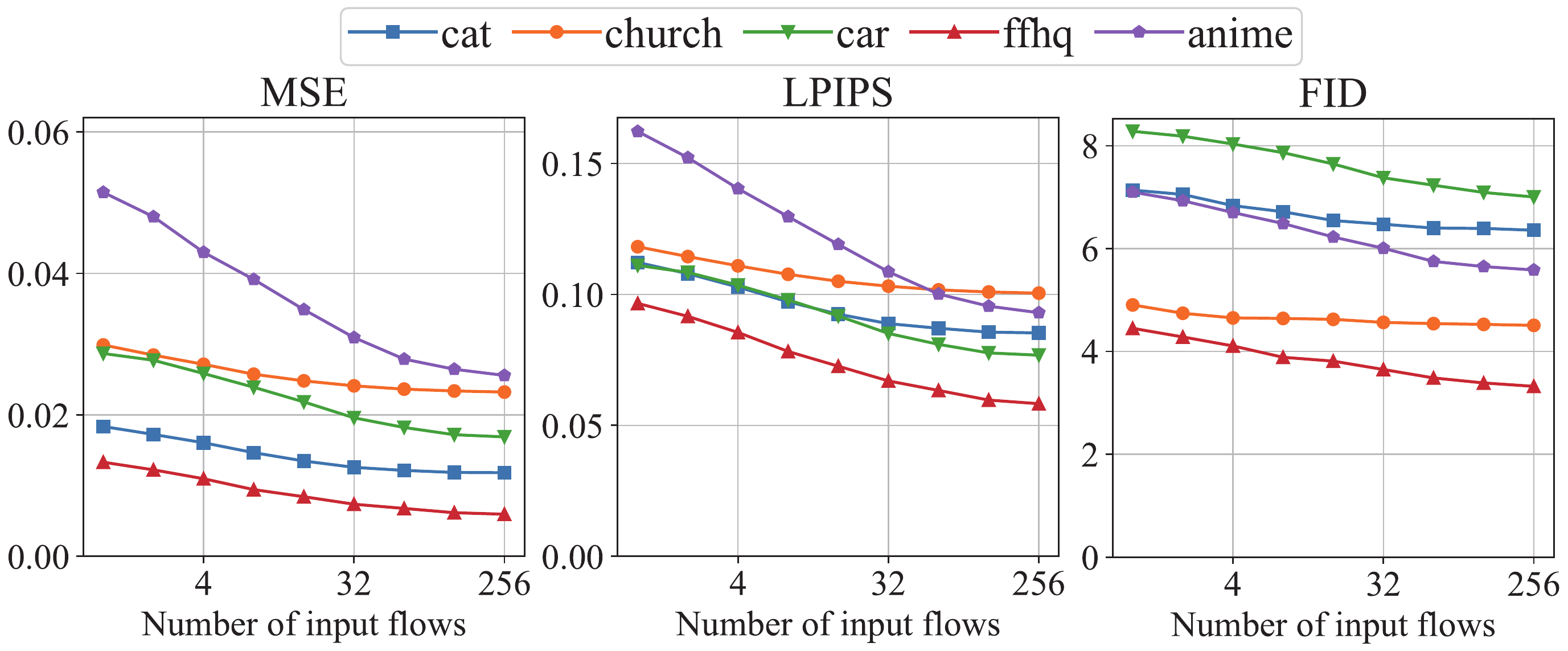}
  \vspace{-4mm}\caption{
  Quantitative comparisons depending on the number of user inputs fed to the latent transformer.
}
  \label{fig:res_flownum}
\end{figure}

\begin{figure*}[t]
  \centering
  \includegraphics*[width=1\linewidth]{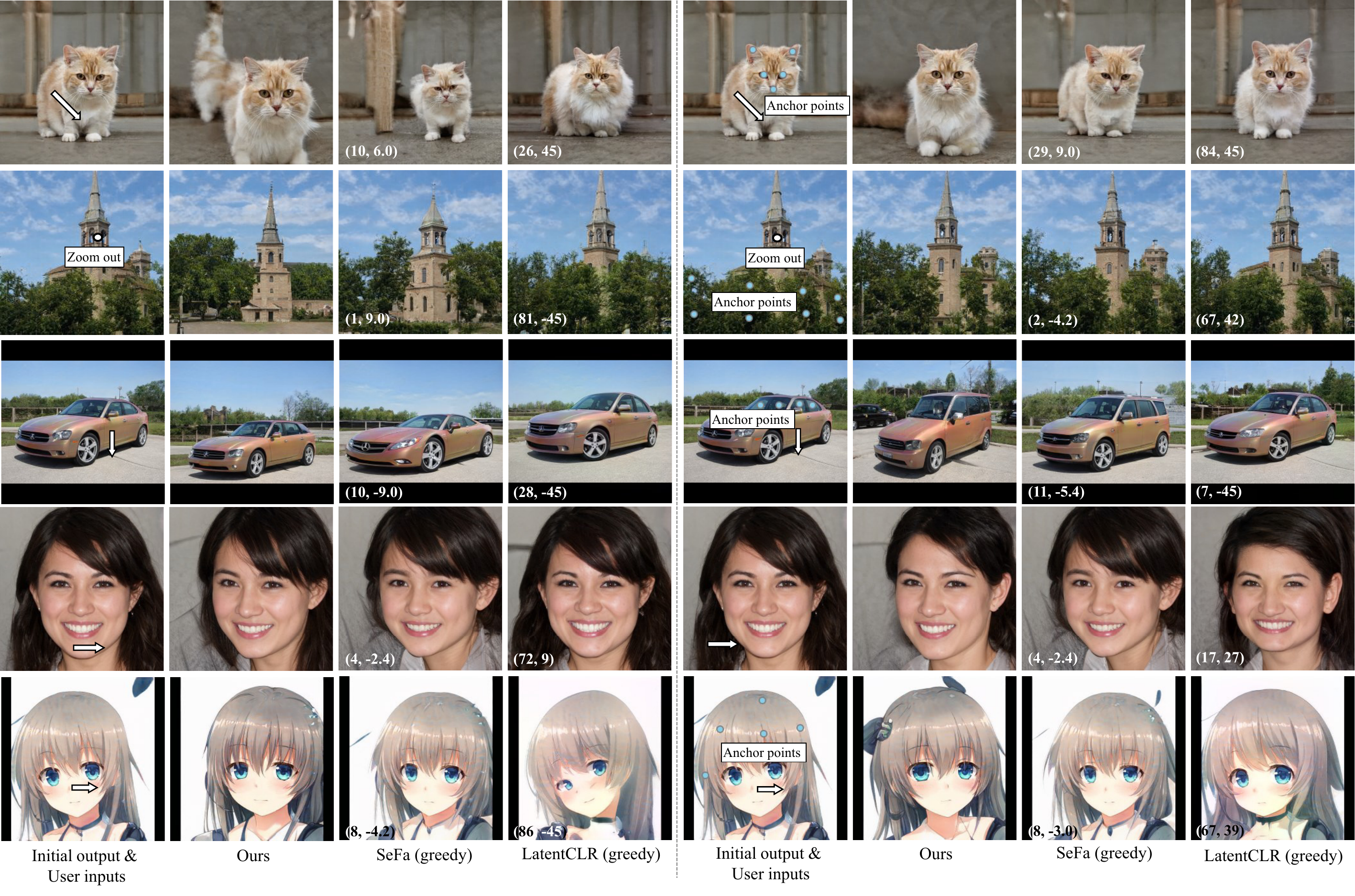}
  \vspace{-4mm}\caption{
Qualitative comparisons between our method, SeFa~\cite{DBLP:journals/corr/abs-2007-06600}, and LatentCLR~\cite{DBLP:conf/iccv/YukselSEY21}. For the user inputs, the white arrows are mouse drags, blue circles are APs, and white circles are zoom operations. 
Two values at the bottom left on each result of SeFa (greedy) and LatentCLR (greedy) show a searched index of a latent direction and the amount of manipulation, respectively. 
}
  \label{fig:res_all}
\end{figure*}

\begin{figure}[t]
  \centering
  \includegraphics*[width=1\linewidth]{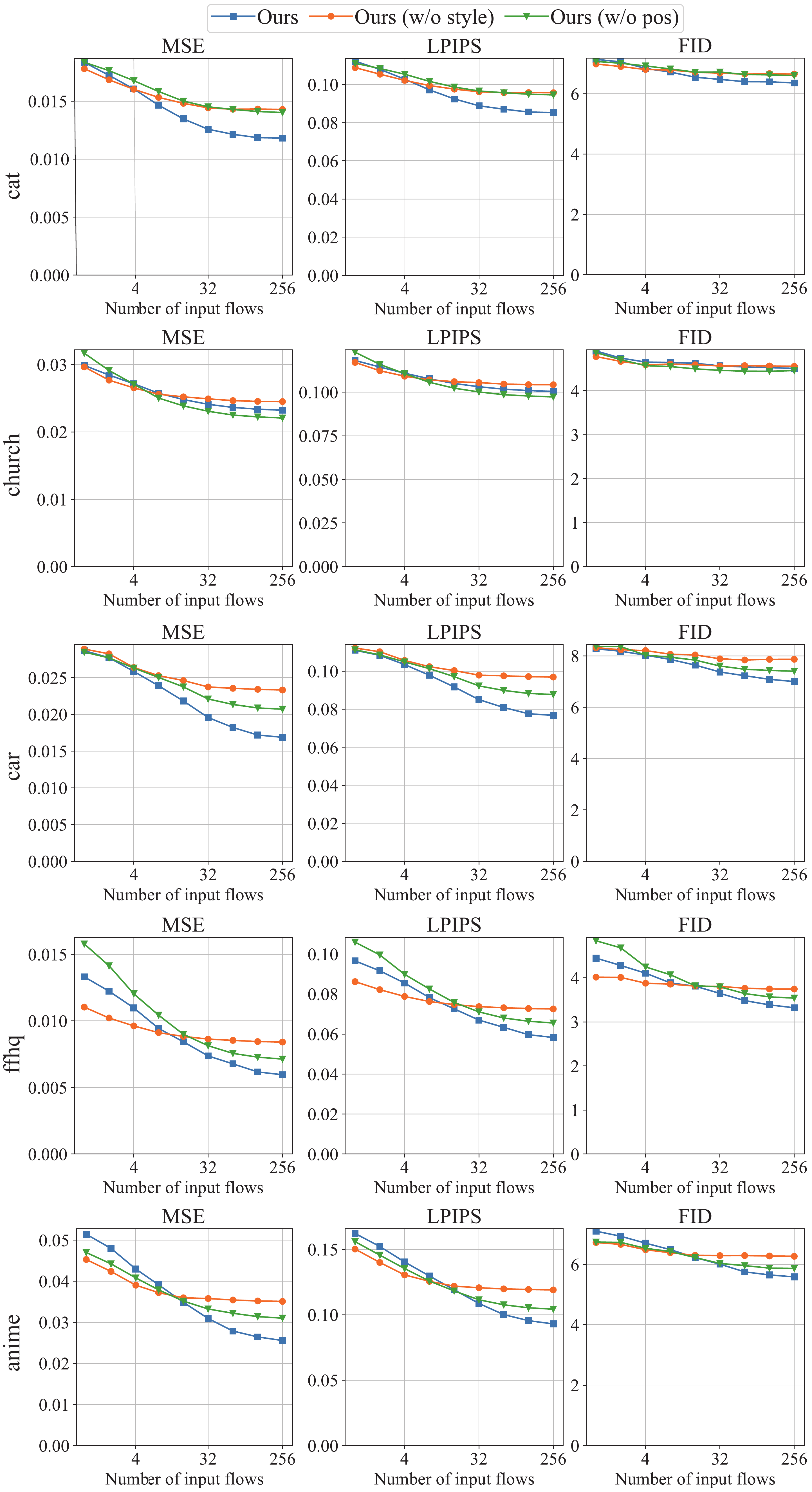}
  \vspace{-4mm}\caption{
Quantitative results of the ablation study, where ``Ours w/o style'' means not using StyleGAN features as input and ``Ours w/o pos'' means not using position embeddings as input. 
From top to bottom, the results are on the cat, church, car, ffhq, and anime datasets. 
}
  \label{fig:ablation}
\end{figure}

\begin{figure}[t]
  \centering
  \includegraphics*[width=1\linewidth]{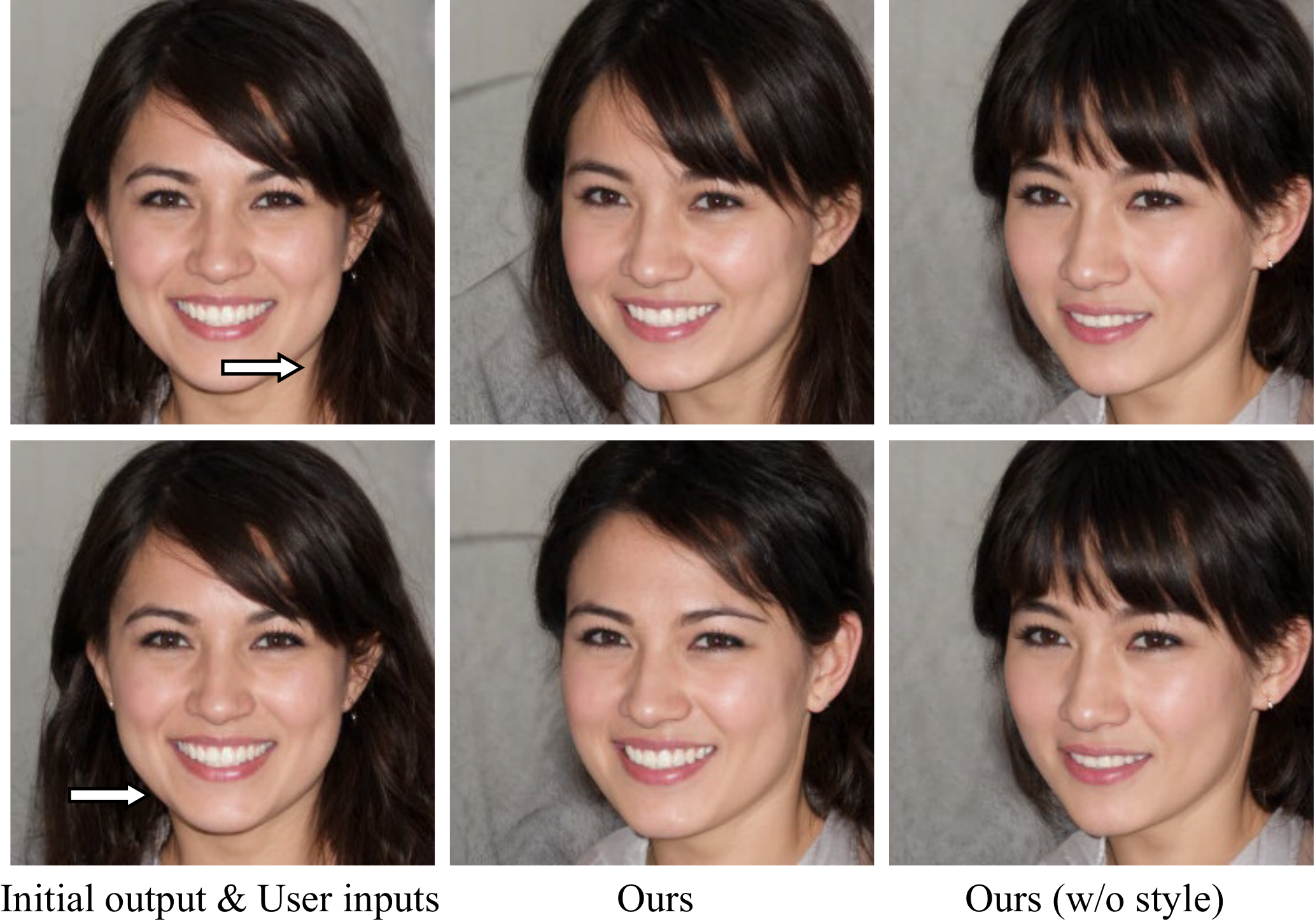}
  \vspace{-4mm}\caption{
Qualitative comparison with (Ours) and without (Ours w/o style) StyleGAN feature maps. 
}
  \label{fig:ablation_ffhq}
\end{figure}

\begin{figure}[t]
  \centering
  \includegraphics*[width=1\linewidth]{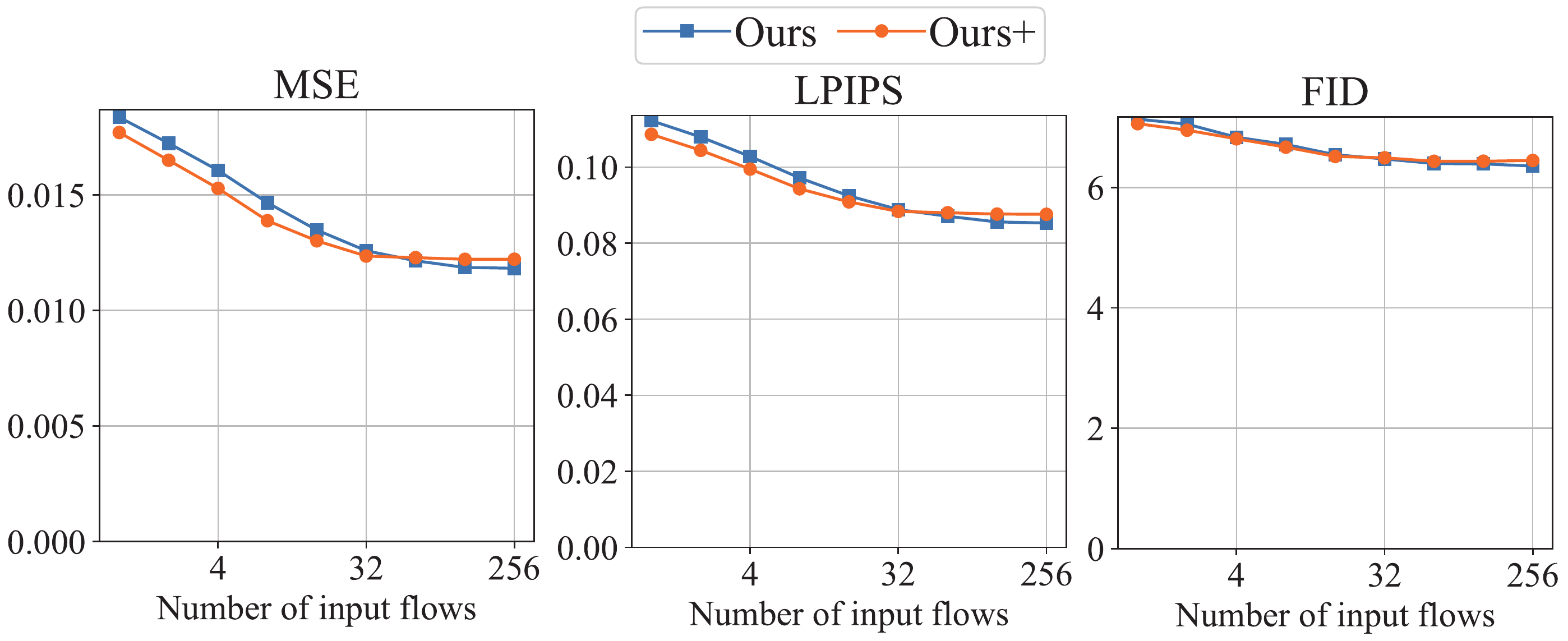}
  \vspace{-4mm}\caption{
\chkA{Influence of the number of flows used for training on the cat dataset. During training, ``Ours'' uses dense flow maps as input, while ``Ours+'' uses subsets of 32 flows randomly sampled from the dense maps. }
}
  \label{fig:ablation_trainnum}
\end{figure}

\paragraph*{Results.} 
As shown in the quantitative results in Table~\ref{tab:quantitative_res}, 
Ours-32, which uses 32 user inputs for our latent transformer, performed best overall. 
SeFa (greedy) is sometimes comparable with Ours-32 but these results were obtained via a greedy parameter search using the ground-truth images. 
SeFa (random) and LatentCLR (random) are significantly worse than the other three methods. 
This means that random search on dozens of parameters is difficult to find appropriate values. 

Nevertheless, it is not so surprising that our method works when many user inputs are used, and thus we analyzed our results with fewer user inputs. 
As shown in Table~\ref{tab:quantitative_res}, because uncertainty for spatial control increases with fewer user inputs, it is reasonable that the results of Ours-1 are worse than those of Ours-32. 
Meanwhile, its results are comparable with SeFa (greedy) and not as bad as those of the other methods. 
Furthermore, we analyzed our results depending on the number of user inputs in Figure~\ref{fig:res_flownum}, where the evaluation metrics gradually increase with fewer user inputs, but their differences are not so significant. 
These results suggest that our method can consider multiple user inputs but generate plausible results even if the number of user inputs is small. 
We verify the effectiveness of our method using a few actual user inputs in the next section. 

\chkA{Note that the difference in our scores between domains (e.g., church is worse than ffhq) may come from the fact that datasets containing more diverse images with complex layouts make latent spaces more entangled during training StyleGAN. }
The examples of the result images in the quantitative evaluation are shown in Appendix~\ref{apend:quantitative}.

\subsection{Qualitative evaluation}
\paragraph*{Evaluation method.} The user inputs used in the quantitative evaluation were the output of the optical flow network and not specified by the user. 
Therefore, we qualitatively evaluated our results using actual user inputs specified with the UI introduced in Section~\ref{sec:interface}. 
Because we have no ground-truth image in this case, we qualitatively validate that our method can control the StyleGAN output in accordance with various user inputs. 
We also confirm whether the existing methods can reproduce the results generated by our method. 
For comparisons, we use the greedy approach assuming that the user finds the best latent directions and parameters. 
To handle larger motions than those in the quantitative evaluation, we set the manipulation ranges to $[-9,+9]$ and $[-45,+45]$ for SeFa and LatentCLR, respectively, and divided these ranges into 31 values for greedy search. 

\paragraph*{Results.} Figure~\ref{fig:res_all} shows several of the qualitative results for each pre-trained StyleGAN. 
We first discuss the results of our method. 
The left side in the top row shows the result of mouse dragging to move the cat's body to the lower right, where the face moves along with the body. 
On the right side, the face position is maintained by additionally specifying the APs on the face. 
The second row shows the results of zooming out the church, where the grass and trees are preserved by specifying additional APs. 
In the third row, by specifying the APs on the top of the car, we can change the height of the car instead of its position. 
In our two results in the fourth row, the user specified the same direction by mouse dragging, but the results differ in accordance with starting points of the annotations. 
Our results in the bottom row show that specifying the APs prevents the head from moving with the anime face. 
Next, we discuss the results of the existing methods, SeFa (greedy) and LatentCLR (greedy). 
The numbers in the lower left of the result images denote the searched indices of the latent directions and the amount of manipulation. 
In many cases, such as the cat in the top row and the anime face in the fifth row, even if the existing methods search for latent directions automatically, they struggle to reproduce similar layouts to ours. 
Although there are several relatively better results, the user actually needs to identify and tune different parameters for each result. 
More results are shown in Appendix~\ref{apend:qualitative} and the supplemental video. 

\subsection{Ablation study}
Figure ~\ref{fig:ablation} shows the quantitative results of the ablation study for our latent transformer.
The evaluation pipeline is the same as that in Section~\ref{sec:eval_quantitative}. 
Here, we compared our original method with the case where StyleGAN feature maps were not used with motion vectors and the case where position embeddings were not used with latent codes. 
The former case verifies whether the semantics of annotated positions are taken into account, and the latter case confirms the effectiveness of the latent space manipulation in the $\mathcal{W}^+$ space. 
In the cat and car results, we can confirm the effectiveness of using StyleGAN feature maps and position embeddings as input.
Although we can observe a similar trend for ffhq and anime when certain numbers of user inputs are given, the evaluation metrics worsen with fewer user inputs. 
These results might come from overfitting caused by increasing the degree of freedom on the inputs. 
Finally, in the church results, there was no significant difference quantitatively. 
Nevertheless, as shown in the results in Figure~\ref{fig:ablation_ffhq},
we can qualitatively observe that the additional feature map input enables us to generate different results in accordance with the starting points of mouse drags, whereas the results are the same if we do not use them as input. 
In addition, the fact that the results differ depending on the positions of the APs in Figures~\ref{fig:teaser}(b) and (c) clearly shows that our method can consider semantics at those positions. 

\chkA{
Furthermore, in Figure~\ref{fig:ablation_trainnum}, we analyzed the influence of the number of flows used for training on the cat dataset. 
Instead of using dense flow maps during training (Ours), we used subsets of 32 flow vectors randomly sampled from the dense maps (Ours+). 
In the results, Ours+ obtained slightly better scores when using 32 or fewer inputs. 
We conducted a similar experiment also on the ffhq dataset and observed the same tendency. 
These analyses show that it would be essential to lessen the gap between training and inference for more accurate prediction. 
}

\begin{figure}[t]
  \centering
  \includegraphics*[width=1\linewidth]{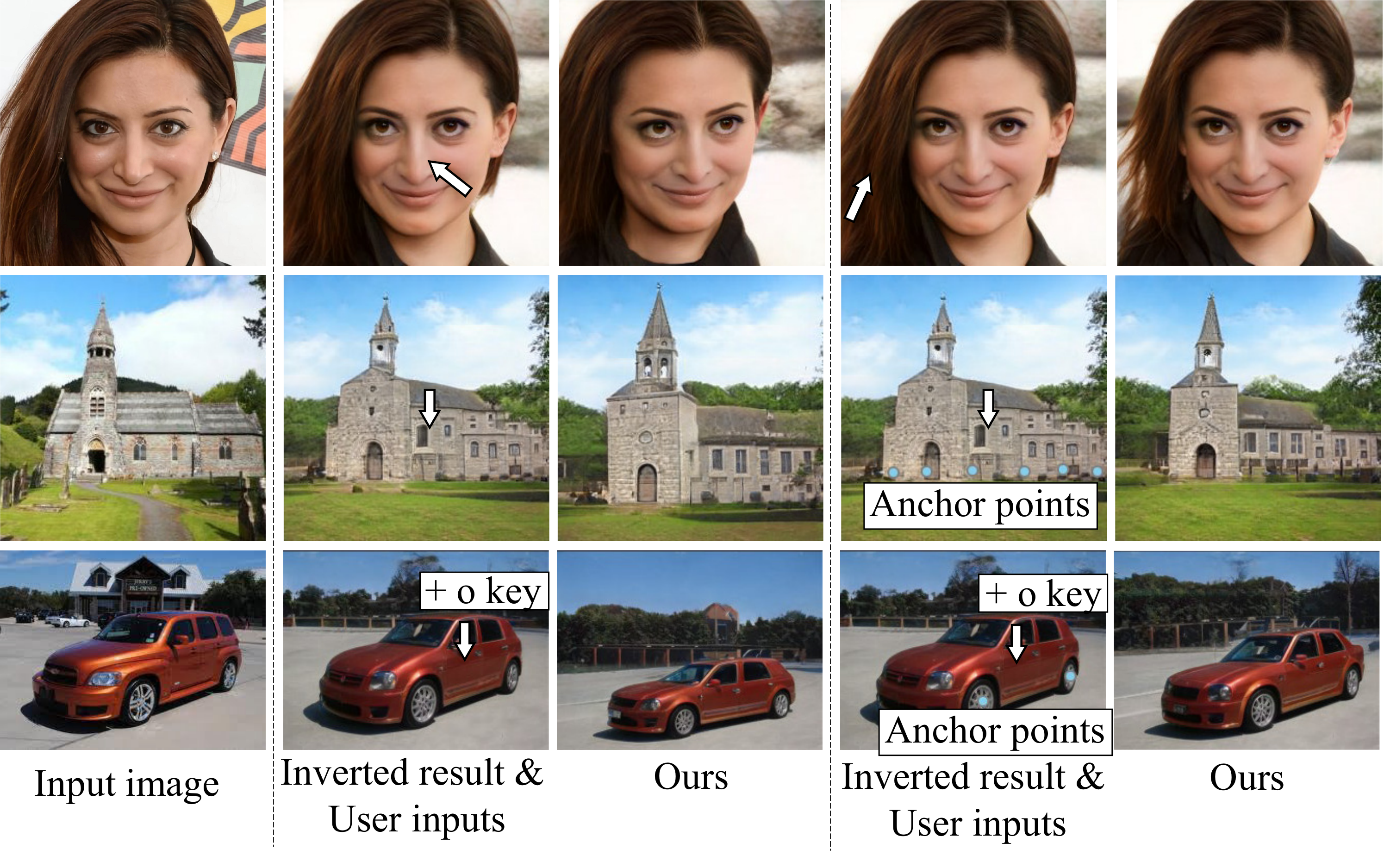}
  \vspace{-4mm}\caption{
Application to real images. As shown in the third and fifth columns, our method can edit the real images (the first column) in accordance with the annotations on the images (second and fourth columns), which we reconstruct from the inverted latent codes. 
}
  \label{fig:inversion}
\end{figure}

\subsection{Application}
\begin{figure}[t]
  \centering
  \includegraphics*[width=1\linewidth]{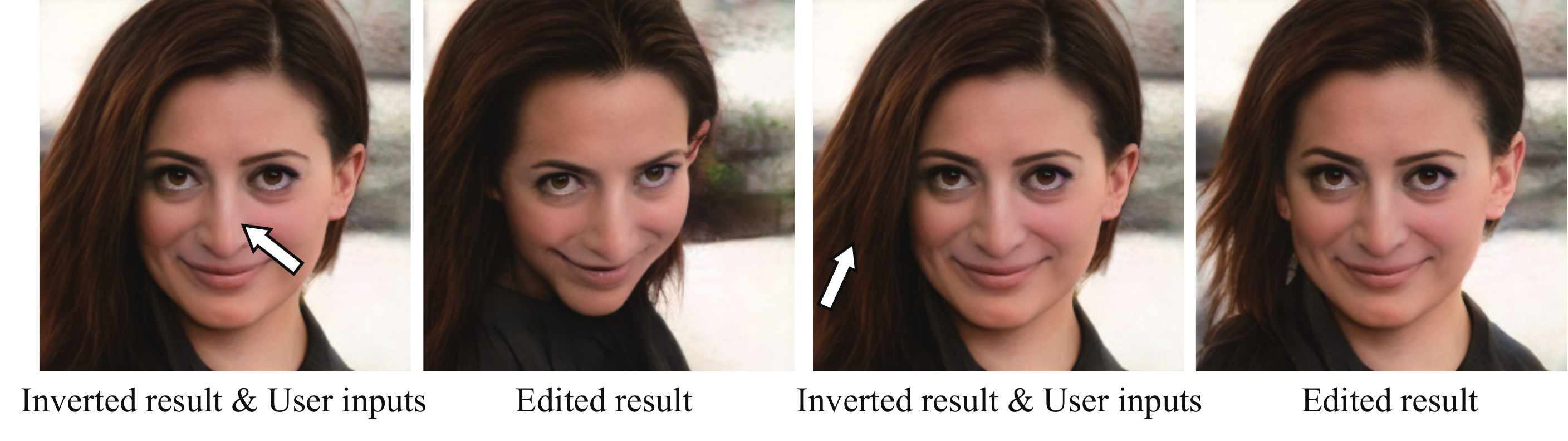}
  \vspace{-4mm}\caption{
  \chkA{Editing real images using our latent transformer trained with latent codes from inverted images in the ffhq dataset. The user inputs are the same as in Figure~\ref{fig:inversion}. }
}
\vspace{-4mm}
  \label{fig:inversion_problem}
\end{figure}

We also verified the editability of our method for real images using GAN inversion. 
We estimated latent codes from real images using ReStyle~\cite{Alaluf_2021_ICCV} and used these latent codes as input to the latent transformer.
Unfortunately, the generalization ability for latent codes derived from real images was low, and artifacts often occurred in the edited results probably because the latent transformer learns not from real data but latent codes from the prior. 
To address this issue, we assumed that operations in latent space do not strongly depend on the absolute values of the latent codes. 
Specifically, instead of the latent codes inverted from real images, we injected the average latent code $\bar{\mathbf{w}}$ into our latent transformer. 
Then, we moved the inverted latent codes toward the obtained latent directions. 
The results in Figure~\ref{fig:inversion} show that our method can manipulate the inverted real images in accordance with the user inputs.
\chkA{
Note that another solution would be to train the latent transformer with latent codes from inverted images to prevent ReStyle codes from being out of the domain for the model. However, as shown in Figure~\ref{fig:inversion_problem}, we frequently observed unnatural results such as distorted faces, which imply that ReStyle codes reside in harder-to-edit regions of the latent space. 
}

\section{Discussion}
\begin{figure}[t]
  \centering
  \includegraphics*[width=1\linewidth]{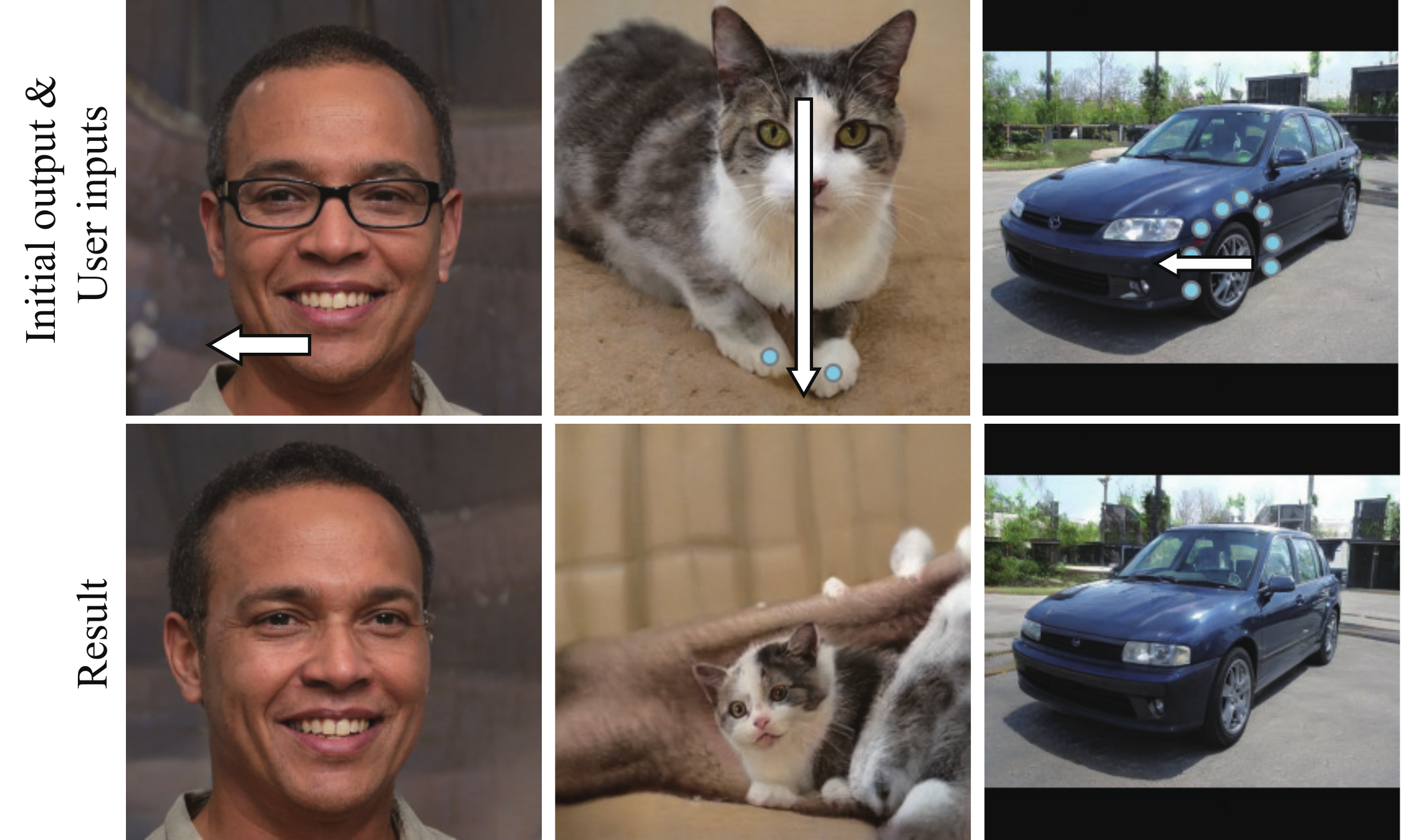}
  \vspace{-4mm}\caption{
\chkA{Failure cases of our method. It sometimes struggles to preserve contents such as glasses and identity (left and middle columns) and perform localized editing such as changing the tire size only (right column). }
}
  \label{fig:failure_cases}
\end{figure}

Although we demonstrated that our latent transformer is user-controllable, it has several limitations. 
First, it is challenging to perform localized editing, such as changing the size of eyes or tires only \chkA{(the right column in Figure~\ref{fig:failure_cases})}. 
\chkA{In addition, our method sometimes struggles to preserve contents such as glasses and identity depending on initial images and extreme user inputs (the left and middle columns in Figure~\ref{fig:failure_cases}). 
}
Although training data are created by randomly manipulating latent codes, this manipulation itself is not disentangled. 
For example, multiple parts of a cat often move simultaneously in the training data, which may also be biased toward particular movements. 
Such data may limit the generalization ability for diverse user inputs. 
To alleviate this problem, we tried to use the $\mathcal{S}$ space~\cite{Wu_2021_CVPR}, which is more disentangled than the $\mathcal{W}^+$ space we used. 
In the $\mathcal{S}$ space, a single channel operation often corresponds to a specific attribute editing. 
We therefore randomly manipulated only one channel in the $\mathcal{S}$ space to create the training data for the latent transformer. 
However, training did not converge well probably because the manipulation of different channels sometimes produced similar flow fields, resulting in one-to-many mapping. 
We leave exploring other disentangled latent spaces for training a latent transformer as future research. 

Another challenge is how to bridge the gap between user annotations specified during the test time and flow fields used during the training time. 
Although our latent transformer can handle variable-length and sparse user inputs, its performance degrades as user inputs decrease due to the ambiguity. 
The future work is to develop a UI to specify dense annotations more efficiently. 
For example, there could be a UI to specify regions where the input flow field is set to zero, effect areas for mouse dragging, or multiple motion vectors. 
There is also room for improvement in 3D annotations using both a mouse and keyboard. 

\section{Conclusion}
In this paper, we have proposed an interactive framework that enables a user to control the layout of StyleGAN images via direct annotations on the image. 
In our framework, we introduced a latent transformer based on a transformer encoder-decoder architecture to handle variable-length user inputs. 
Using synthetic images and pseudo-user inputs, our method can train the latent transformer, without manual supervision. 
Evaluation experiments showed that our method can manipulate latent codes in accordance with user inputs specified with our UI.  
We also demonstrated the effectiveness of our method through quantitative and qualitative comparisons with existing methods. 
Although we still have challenging future works as previously mentioned, we believe that our work, which is the first approach to the user-controllable latent code transformation, will inspire successive work for intuitive latent space manipulation for GANs. 
\chkA{
Integrating more advanced GANs (e.g.,  StyleNeRF~\cite{gu2021stylenerf} and StyleGAN3~\cite{DBLP:conf/nips/KarrasALHHLA21}) into our framework is also an interesting future direction. 
}

\begin{figure*}[t]
  \centering
  \includegraphics*[width=1\linewidth]{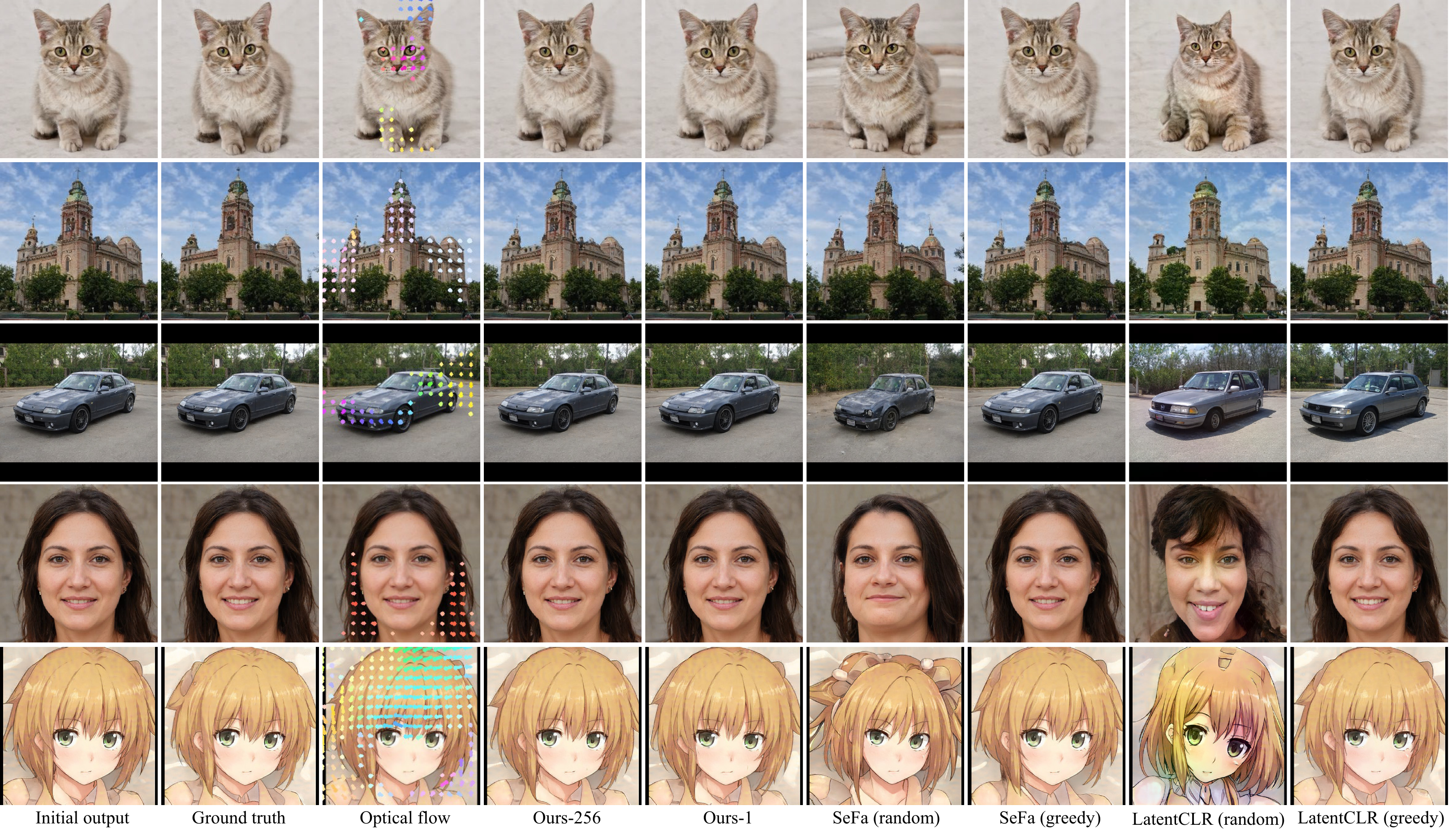}
  \vspace{-4mm}\caption{
  Example images used in the quantitative evaluation. From left to right, initial StyleGAN output images, ground-truth images, optical flow, and result images for each method. 
}
  \label{fig:res_quantitative}
\end{figure*}

\begin{figure*}[t]
  \centering
  \includegraphics*[width=1\linewidth]{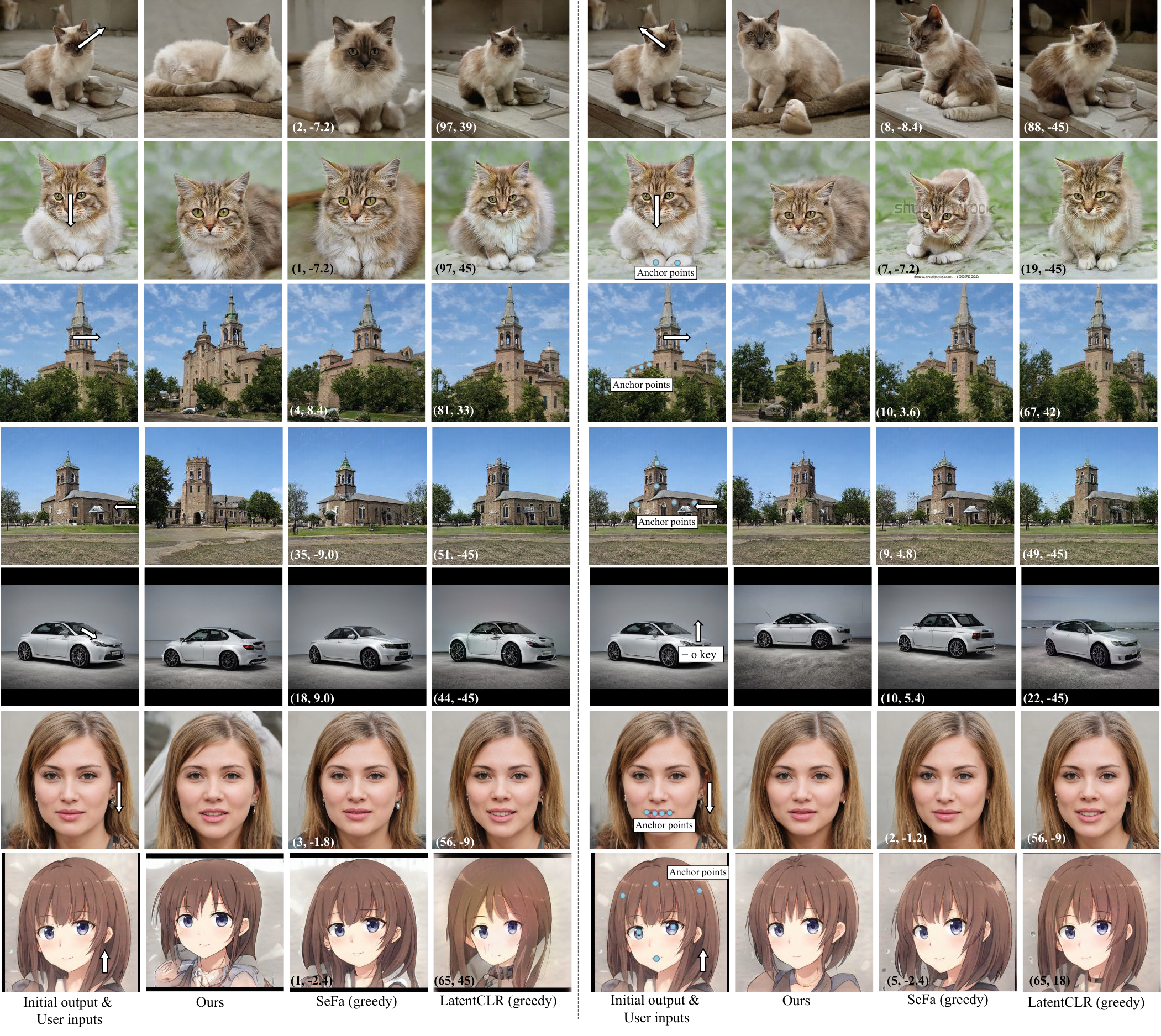}
  \vspace{-4mm}\caption{
  Additional comparisons on the cat, church, \chkA{car, ffhq, and anime} datasets. 
}
  \label{fig:res_appendix}
\end{figure*}

\section*{Acknowledgements}
We thank Prof. Yoshihiro Kanamori for his fruitful discussions. We also thank the anonymous reviewers for their constructive feedback. This work was supported by JSPS KAKENHI Grant Number~20K19816. 


\printbibliography        

\appendix

\section{
Result Images in Quantitative Evaluation}
\label{apend:quantitative}
Figure~\ref{fig:res_quantitative} shows examples of the result images used for the quantitative evaluation. 
From the first to the third column, we show the initial output image from $\mathbf{w}_{before}$, the ground-truth image from $\mathbf{w}_{after}$, and the corresponding forward flow superimposed on the initial output image, respectively. 
From the fourth column, we show the results of each method.
Because the optical flow model often fails to estimate flows of large motions, the synthetic data used in the quantitative evaluation are limited to small motions. 
As shown in the figure, the results of SeFa (random) and LatentCLR (random) have many artifacts and are significantly different from the ground-truth image.  
For the other methods, while it may be difficult to discern qualitative differences between them in these datasets, we validated the effectiveness of our method quantitatively, as already demonstrated in the paper.

\section{Additional Qualitative Results}
\label{apend:qualitative}
\chkA{Figure~\ref{fig:res_appendix} shows} the additional qualitative comparisons, where our results were manually generated with our UI. 
Our results demonstrate that our method can control spatial layout changes in accordance with user inputs. 
However, it is not easy for SeFa (greedy) and LatentCLR (greedy) to reproduce our results. In addition, these methods need to identify appropriate latent directions and tune their parameters. 

\end{document}